\documentclass{article}

\usepackage{microtype}
\usepackage{graphicx}
\usepackage{booktabs}
\usepackage{caption}
\usepackage[backref=page]{hyperref}

\usepackage{preprint}

\usepackage{amsmath}
\usepackage{amssymb}
\usepackage{mathtools}
\usepackage{amsthm}
\usepackage{bbm}
\usepackage[capitalize,noabbrev]{cleveref}
\usepackage{balance}
\usepackage{enumitem}
\usepackage{stfloats}

\renewcommand{\paragraph}[1]{\textbf{#1}}

\renewcommand*{\backref}[1]{}
\renewcommand*{\backrefalt}[4]{%
    \ifcase #1\or Cited on page~#2.\else Cited on pages~#2.\fi%
}

\icmltitlerunning{Temporal Task Diversity: Inductive Biases Under Non-Stationarity in Synthetic Sequence Modelling}

\icmlkeywords{Science of deep learning, inductive bias, continual learning, non-stationary learning, transformers, in-context linear regression}

\date{May 18, 2026}

\begin{document}

\twocolumn[
\icmltitle{Temporal Task Diversity:\texorpdfstring{\\}{ }Inductive Biases Under Non-Stationarity in Synthetic Sequence Modelling}

\icmlsetsymbol{equal}{*}

\begin{icmlauthorlist}
\icmlauthor{Afiq Abdillah Effiezal Aswadi}{equal,indie}
\icmlauthor{Oliver Britton}{equal,ox}
\icmlauthor{Ross Baker}{equal,ox}
\icmlauthor{Matthew Farrugia-Roberts}{ox}
\end{icmlauthorlist}

\icmlaffiliation{indie}{Independent}
\icmlaffiliation{ox}{University of Oxford}

\icmlcorrespondingauthor{MFR}{matthew@far.in.net}

\vskip 0.3in
]

\printAffiliationsAndNotice{\textsuperscript{*}Equal contribution (random order). Cite as: Effiezal Aswadi et~al., 2026.}

\begin{abstract}
Modern deep learning science often assumes that neural networks learn from a fixed data distribution. However, many practically important learning problems involve data distributions that change throughout training. How does such \emph{non-stationarity} impact the inductive biases of deep learning towards models with different structural, generalisation, and safety properties? A fruitful testbed for studying inductive bias is in-context linear regression sequence modelling, where small transformers display strikingly different generalisation patterns depending on the diversity of the (fixed) training task distribution.
In this paper, we explore the effect of diversifying the task distribution \emph{across training time,} finding that such temporal diversity leads to an increased bias towards generalisation over memorisation.
\end{abstract}

\section{Introduction}
\label{sec:intro}

The science of deep learning aims to understand phenomena exhibited by deep neural networks.
As an example phenomenon, transformers trained on linear regression data learn different in-context learning algorithms depending on how many latent regression vectors (or \emph{tasks}) they encounter during training \citep[the \emph{task diversity};][]{Raventos+2023}:
\begin{itemize}
    \item
        With low task diversity, the transformer learns a specialised algorithm for distinguishing between the specific latent tasks encountered during training.
    \item 
        With high task diversity, the transformer learns a general algorithm resembling ridge regression that accommodates both training tasks and novel tasks.
\end{itemize}
This \emph{task diversity threshold} phenomenon, and other phenomena in similar synthetic sequence modelling settings, have enabled us to study deep learning's so-called \emph{inductive biases}.
For example, \citet{Hoogland+2025} and \citet{Carroll+2025} cast these phenomena as neural networks navigating a trade-off between training loss and model complexity.

Prior work in this vein has considered independently and identically distributed sequence data.
However, many important learning problems involve data that is \emph{non-identically} distributed.
For example, modern foundation models undergo multiple stages of pre- and post-training, all with different data distributions \citep{Ouyang+2022}.
Moreover, curriculum learning methods deliberately alter the data distribution throughout training \citep{Bengio+2009}, and on-policy (multi-agent) reinforcement learning algorithms generate experience data using the latest policies \citep{Sutton+Whitehead1993,Papoudakis+2019}.
Finally, due to the dynamic nature of our world, many data generating processes encountered in practice evolve over time \citep{Clements+Henry1999,Milly+2008,Nestor+2019}.

The introduction of non-stationarity has the potential to alter the way deep learning selects between models with different structures and behaviours. Understanding the principles governing this selection is key to ensuring the safety of future advanced learning systems
    \citep{Wentworth2021,PepinLehalleur+2025}.
With this motivation in mind, we offer the following contributions:
\begin{itemize}
    \item
        We extend the synthetic sequence modelling problem of \citet{Raventos+2023} to allow for a dynamically varying distribution of latent tasks (\cref{sec:setting}).
    \item 
        We show experimentally that increasing the rate of change of the task distribution decreases the task diversity threshold (\cref{sec:experiments}), below which the transformer tracks the changing set of latent tasks (\cref{sec:experiments:pursuit}).
    \item
        We discuss possible explanations of this phenomenon, including that learning is biased towards both model simplicity and, newly, model \emph{stability} (\cref{sec:discussion}).
\end{itemize}
Our results suggest that non-stationary deep learning exhibits an inductive bias of a somewhat richer nature than that exhibited in stationary learning, motivating further study into the extent and mechanism of this phenomenon in this and similar deep learning settings.

\section{Related work}
\label{sec:relwk}

\paragraph{Inductive biases in synthetic sequence modelling.}
Machine learning problems are often underspecified, meaning that the provided data do not identify a unique solution model \citep[\S8]{Breiman2001}.
Inductive biases refer to the combined effects of the architecture and learning algorithm in determining the model chosen by a learning process \citep{Mitchell1980}.
In modern deep learning, transformers trained on synthetic sequence modelling data \citep{Garg+2022} have furnished striking examples of critical thresholds at which the solutions found by the learning process change.
Examples of such phenomena include the emergence of in-context learning given data sets with certain properties
    \citep{Chan+2022},
or a transition between specialised and generic in-context learning algorithms with increasing task diversity
    \citep{Raventos+2023,He+2024,Park+2025}.
To the best of our knowledge, no prior work has studied the inductive biases of \emph{non-stationary} synthetic sequence modelling.

\paragraph{Learning from non-stationary data.}
Non-stationarity is core to the problems studied in the literatures on
    lifelong/continual learning \citep{Schlimmer+Fisher1986,Thrun+Mitchell1995,Chen+Liu2018,Parisi+2019,Wang+2024};
    concept drift or non-stationary learning \citep{Widmer+Kubat1996,Tsymbal2004,Ditzler+2015};
and
    reinforcement learning \citep{Sutton+Whitehead1993,Sutton+Barto2018,Papoudakis+2019,Igl+2021}.
Note that most prior work in these settings aims to engineer architectures or learning algorithms pursuant to various learning objectives.
In this work, we instead aim to advance scientific understanding of the phenomena that non-stationarity induces in mainstream architectures and learning algorithms---comparable to work studying phenomena like catastrophic forgetting
    \citep{McCloskey+Cohen1989,Ratcliff1990,French1999,Goodfellow+2014}
or hysteresis
    \citep{Igl+2021,Nikishin+2022}.

\section{Problem setting}
\label{sec:setting}
We study transformers trained on synthetic in-context linear regression data, extending the setting of \citet{Raventos+2023} to non-stationary task distributions. 

\paragraph{Tasks and sequences.}
A \emph{task} is a latent regression vector $\mathbf t \in \mathbb R^D$.
Given a task $\mathbf t$, define a conditional distribution $q(S \mid \mathbf t)$ over sequences $S = (x_1, y_1, \ldots, x_K, y_K) \in (\mathbb R^{D} \times \mathbb R)^K$ by independently sampling inputs $x_k \sim \mathcal N(0, I_D)$ and setting $y_k = \mathbf t^\top x_k + \varepsilon_k$ with $\varepsilon_k \sim \mathcal N(0, \sigma^2)$.
Define an unconditional sequence distribution $q(S) = \int q (S \mid \mathbf t) q(\mathbf t) \text d \mathbf t$ by first sampling a task from a \emph{task distribution} $q(\mathbf t)$ and then sampling a sequence from the corresponding conditional distribution.

Following \citet{Raventos+2023}, we parametrise a family of task distributions $q_M(\mathbf t)$ by a \emph{task diversity} $M \in \mathbb N \cup \{\infty\}$. A task set $\mathcal T_M = \{\mathbf t_1, \ldots, \mathbf t_M\}$ induces the discrete uniform task distribution $q_M(\mathbf t) = \text{Uniform}(\mathcal T_M)$. We also define $q_\infty(\mathbf t) = \mathcal N(0, I_D)$.

\paragraph{Non-stationary task distributions.}
In the stationary setting of \citet{Raventos+2023}, the task set $\mathcal T_M$ is fixed throughout training. We generalise this by allowing the task set to vary with training step $\tau \in \{1, \ldots, T\}$, writing $\mathcal T_M(\tau) = \{\mathbf t_1(\tau), \ldots, \mathbf t_M(\tau)\}$ for the task set at step $\tau$ and $q_M^{(\tau)}(\mathbf t) = \text{Uniform}(\mathcal T_M(\tau))$ for the corresponding task distribution.

\paragraph{Mean squared error objective.}
Given a sequence $S$, write $S_{\le k} = (x_1, y_1, \ldots, x_{k-1}, y_{k-1}, x_k)$ for the \emph{context} when predicting $y_k$. A predictor $f$ maps contexts to predicted labels. Given a data distribution $q(S)$, define the population loss
\begin{equation}
\ell(f) = \mathop{\mathbb{E}}\limits_{S \sim q} \left[ \frac 1 K \sum^K_{k=1} (f(S_{\le k}) - y_k)^2 \right].
\end{equation}
For a transformer with parameters $w$, we write $\ell(w)$ for the loss of $f(\cdot; w)$, and $\ell^M(w)$ or $\ell^{M, \tau}(w)$ when the data distribution is $q_M$ or $q_M^{(\tau)}$ respectively.

\paragraph{Optimal predictors.}
\citet{Raventos+2023} showed that the optimal estimator for the $k$-th prediction, minimising the $k$-th term in $\ell (f)$, is the Bayesian posterior mean $\hat y_k = \hat{\mathbf t}_k^\top x_k$ where $\hat{\mathbf t}_k = \mathbb E[\mathbf t \mid S_{\le k}]$, with the expectation taken over the task distribution $q(\mathbf t)$ as the prior.

For finite task diversity (prior $q_M(\mathbf t)$), the posterior given context $X = [x_1\ \cdots\ x_{k-1}]^\top$ and $\mathbf y= (y_1, \ldots, y_{k-1})$ is
\begin{equation}
p(\mathbf t \mid X, \mathbf y) = \frac{\exp\left(-\frac{1}{2\sigma^2} ||\mathbf y - X \mathbf t||^2\right)}{\sum^M_{m=1} \exp\left(-\frac{1}{2\sigma^2} ||\mathbf y - X \mathbf t_m||^2\right)}
\end{equation}
for $\mathbf t \in \mathcal T_M$ (and $0$ otherwise). The posterior mean then yields the \emph{discrete minimum mean squared error} (dMMSE) predictor, with task estimate
\begin{equation}
\hat{\mathbf t}_k^M = \frac{\sum^M_{m=1} \exp\left( -\frac{1}{2\sigma^2} \sum^{k-1}_{j=1} (y_j - \mathbf t_m^\top x_j)^2 \right) \mathbf t_m}{\sum^M_{m=1} \exp (-\frac{1}{2\sigma^2}\sum^{k-1}_{j=1} (y_j - \mathbf t_m^\top x_j)^2)}.
\end{equation}

For infinite task diversity (prior $q_\infty(\mathbf t)$), the posterior is a Gaussian with parameters
\begin{equation}
\begin{aligned}
\mathbf t \mid X, \mathbf y \sim & \;\mathcal N(\mu_k, \Sigma_k), \\
\Sigma_k = \left(\frac 1 {\sigma^2} X^\top X + I_D \right)^{-1}&, \quad
\mu_k = \frac{1}{\sigma^2} \Sigma_k X^\top \mathbf y.
\end{aligned}
\end{equation}
The posterior mean yields the \emph{ridge} predictor, with task estimate
\begin{equation}
\hat{\mathbf t}_k^\infty = (X^\top X + \sigma^2 I_D)^{-1} X^\top \mathbf y
\end{equation}
Thus dMMSE is optimal under $q_M$ for finite $M$, but is specialised to the particular tasks in $\mathcal T_M$, while ridge is optimal only under $q_\infty$ and generalises to any task drawn from $\mathcal N(0, I_D)$.

\section{Reducing the task diversity threshold}
\label{sec:experiments}

\begin{figure*}
  \centering
  \includegraphics[width=\textwidth]{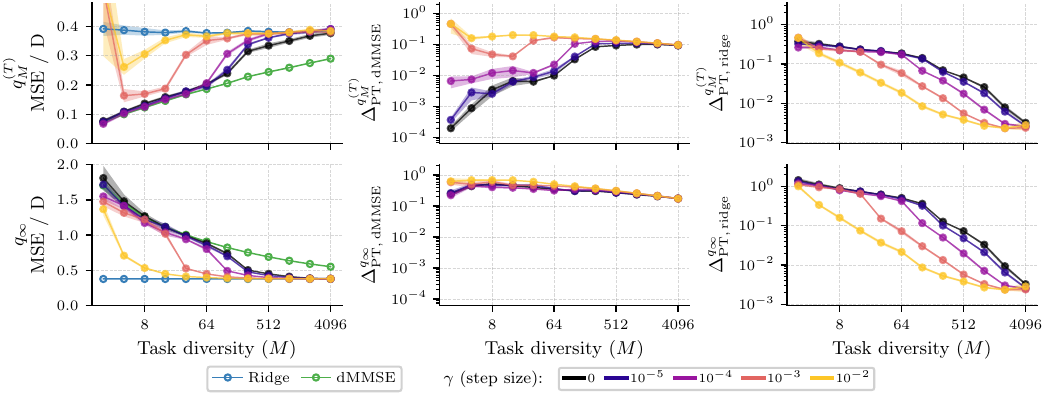}
  \caption{\textbf{Increasing non-stationarity via MALA random walk shifts the dMMSE-ridge transition to lower task diversities.} We show results using the final $M$ tasks at the end of pretraining (\textit{top row}) and on new tasks drawn from $\mathcal T_\text{True} = \mathcal N(0, I_D)$ (\emph{bottom row}), for transformers trained via tasks updating according to a MALA random walk with step size $\gamma$ at each step of training. We vary $\gamma$ from $0$ (dark) to $10^{-2}$ (light). The \textit{left column} compares the normalised loss of pretrained transformers (PTs) with increasing static task diversity $M$ to that of the dMMSE and ridge reference predictors. The \textit{middle} and \textit{right columns} show the mean squared distance $\Delta_\text{PT,dMMSE}$ and $\Delta_\text{PT,Ridge}$ between the PT's predictions and those of dMMSE and ridge respectively. Results averaged across 7 seeds, shaded region shows $\pm 1$ standard error. Five outlier runs excluded due to training instability (see \cref{app:unstable}).
  At low step sizes, the PT reproduces the results from the stationary setting in \citet{Raventos+2023}: it approximates dMMSE at low $M$ and transitions to ridge at high $M$. As $\gamma$ increases, this transition shifts towards lower task diversities. At extremely low task diversity and large step size, training is unstable.}
  \label{fig:mala}
\end{figure*}

In this section, we show that increasing the rate of change of the task distribution lowers the task diversity threshold---transformers learning from a more rapidly varying data distribution increasingly prefer the ridge solution.
We study two forms of non-stationarity: in \cref{sec:experiments:mala}, we change the distribution by a small amount every step, and in \cref{sec:experiments:dirichlet} we resample tasks at randomly chosen times.

Following \citet{Raventos+2023}, we train 8-layer transformers using the problem setting described in \cref{sec:setting} (see \cref{app:experiment-details} for details).
After training, we compare the predictions of the pretrained transformers (PT) to those of the optimal predictors (dMMSE or ridge) by computing the mean square prediction differences
\begin{equation}\label{eq:deltas}
\Delta^q_\text{PT,\texttt{ref}} = \mathop{\mathbb{E}}\limits_{S \sim q} \left[ \frac{1}{KD} \sum^K_{k=1} \left(f(S_{\le k}; w) - \hat y_k^\texttt{ref} \right)^2 \right]
\end{equation}
where $\texttt{ref}$ is either dMMSE or ridge and $q$ is either in-distribution sequences ($q_M^{(\tau)}$) or out-of-distribution sequences ($q_\infty$). These metrics serve as estimates of the corresponding $L_2$ distances in function space.

\subsection{Random walk non-stationarity}
\label{sec:experiments:mala}

First, we introduce non-stationarity by evolving each task vector independently via the Metropolis-Adjusted Langevin Algorithm \citep[MALA;][]{Roberts+Tweedie1996}, a Markov chain Monte Carlo method that targets $\mathcal N(0, I_D)$ as its stationary distribution. We initialise $\mathbf t_m(0) \sim \mathcal N(0, I_D)$. At each training step $\tau$, each task $\mathbf t_m(\tau)$ is updated by first computing a proposal
\begin{equation}
\tilde{\mathbf t}_m(\tau + 1) = \left(1 - \frac \gamma 2\right)\mathbf t_m(\tau) + \sqrt \gamma \; \boldsymbol \xi_m(\tau)
\end{equation}
where $\boldsymbol \xi_m(\tau) \sim \mathcal N(0, I_D)$ and $\gamma > 0$ is the step size. The proposal is then accepted or rejected via a Metropolis-Hastings step, where $\mathbf t_m(\tau + 1) = \tilde{\mathbf t}_m$ with probability $\alpha$ and $\mathbf t_m(\tau + 1) = \mathbf t_m(\tau)$ otherwise, where
\begin{equation}
    \alpha = \min\left(1, \exp\left[\frac{\gamma}{8}(||\mathbf t_m(\tau)||^2 - ||\tilde{\mathbf t}_m||^2)\right]\right).
\end{equation}
This ensures that $\mathcal N(0, I_D)$ is the stationary distribution of the chain so that the marginal distribution of each task at every training step is $\mathcal N(0, I_D)$ regardless of $\gamma$. The magnitude of the step size controls the rate of change of the task distribution, so that small $\gamma$ approximates the stationary setting with the task distribution changing only slightly between training steps, while large $\gamma$ causes the task set to change quickly over the course of training.

We train a separate model for each combination of static task diversity $M \in \{2, 4, 8, \ldots, 4096\}$ and MALA step size $\gamma \in \{10^{-5}, 10^{-4}, 10^{-3}, 10^{-2}\}$. We also tested larger step sizes ($\gamma \ge 10^{-1}$), but found that training was unstable and the transformer converged to neither dMMSE nor ridge.

The results are shown in \cref{fig:mala}. At step size~0, we recover the stationary results of \citet{Raventos+2023}, where varying the task diversity causes the transformers to transition from approximating dMMSE at low values of $M$ to approximating ridge at high values of $M$. As we introduce non-stationarity by increasing $\gamma$, this transition occurs at lower task diversities. For the lowest task diversity and largest step sizes, training is unstable (see \cref{app:unstable}).

\begin{figure*}
  \centering
  \includegraphics[width=\textwidth]{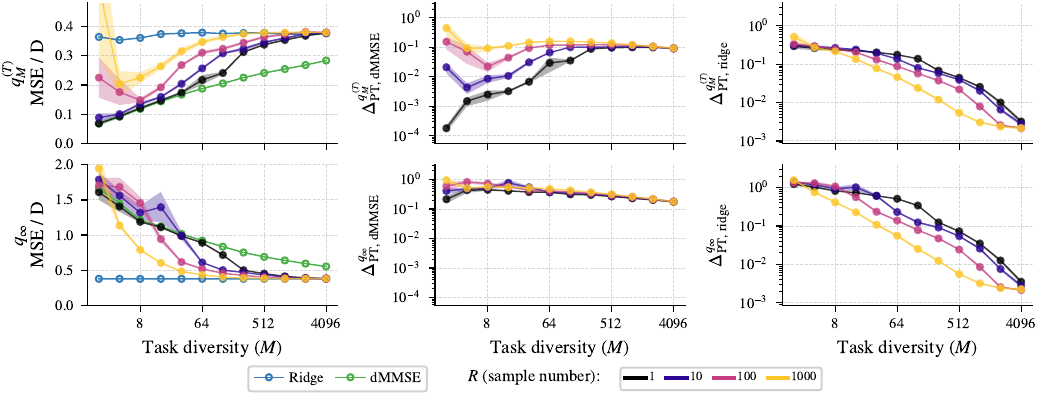}
  \caption{\textbf{Increasing non-stationarity via resampling shifts the dMMSE-ridge transition to lower task diversities.} As in \cref{fig:mala}, we show results for both the final $M$ tasks at the end of pretraining (\textit{top row}) and for new tasks drawn from $\mathcal T_\text{True} = \mathcal N(0, I_D)$ (\emph{bottom row}). We use transformers pretrained with resampling non-stationarity, for which we vary the sample number $R$ from $1$ (dark) to $1000$ (light). The \textit{left column} compares the normalised loss of the pretrained transformers (PTs) to that of the dMMSE and ridge reference predictors. The \textit{middle} and \textit{right columns} show the mean squared distances $\Delta_\text{PT,dMMSE}$ and $\Delta_\text{PT,Ridge}$ between the predictions of PT and those of dMMSE and ridge respectively. Note that these dMMSE and ridge predictions do not depend on $R$.
  Results averaged across 7 seeds, shaded regions show $\pm 1$ standard error. Two outlier runs excluded due to training instability (see \cref{app:unstable}).
  We see again that the transition from dMMSE to ridge shifts to lower task diversities as the sample number $R$ increases.}
  \label{fig:dirichlet}
\end{figure*}

\subsection{Resampling non-stationarity}
\label{sec:experiments:dirichlet}

We also test the effect of gradually updating the task set $\mathcal T_M(\tau) = \{\mathbf t_1(\tau), \ldots, \mathbf t_M(\tau)\}$ throughout training by resampling each task at $R-1$ randomly selected training steps. In particular, each task is given by a stepwise function 
\begin{equation}
\mathbf{t}_{i}(\tau) = \sum_{j=0}^{R-1} \mathbbm{1}_{[\tau_{i,j}, \tau_{i,j+1})} (\tau) \mathbf{t}_{i,j} \ \ \ \forall i \in \{ 1,...,M \}
\end{equation}
where $\mathbf{t}_{i,j} \overset{\text{i.i.d.}}{\sim} \mathcal N(0, I_D)$. Each initial sample happens at training step $\tau_{i,0} = 1$, and the updates occur at steps
\begin{equation}
\tau_{i,j} = T \sum_{k=1}^{j} \delta_{i,k} \ \ \ \forall j \in \{ 1,...,R-1 \}
\end{equation}
where $(\delta_{i,1}, \ldots, \delta_{i,R}) \overset{\text{i.i.d.}}{\sim} \text{Dir}(\mathbf{1}_{R})$. Here we use a Dirichlet distribution since it has the property that $\sum_{j=1}^{R} \delta_{i,j} = 1$, allowing each $\delta_{i,j}$ to represent a proportion of the training run. Finally we let $\tau_{i,R} = T+1$ so that $\mathbf{t}_{i}(\tau)$ is defined up to and including the final training step.

The sample number $R$ controls how often the tasks are resampled, with $R=1$ corresponding to the stationary setting and large $R$ corresponding to frequent task set updates. We designed this setup so that the tasks are updated asynchronously and at irregular intervals, imitating how a data distribution might naturally vary over time. Note that the model sees a total of $MR$ realised task vectors over the course of training. 

We train a separate model for each combination of task diversity $M \in \{2, 4, 8, \ldots, 4096 \}$ and sample number $R \in \{1, 10, 100, 1000\}$. The results are shown in \cref{fig:dirichlet}. As usual, we see transformers trained with low $M$ approximating dMMSE, while those trained with high $M$ approximate ridge. For $R=1$ we recover the stationary results of \citet{Raventos+2023}, and as $R$ increases the transition between dMMSE and ridge shifts to lower task diversities.
Once again, for the lowest task diversities and largest sample numbers, training is unstable.

\section{Continual specialisation}
\label{sec:experiments:pursuit}

\Cref{sec:experiments} shows the effect of increasing non-stationarity on whether the \emph{fully-trained} transformer approximates dMMSE or ridge. In this section, we study the evolution of the transformer's behaviour \emph{throughout training} from multiple perspectives.
We find that those transformers that eventually approximate dMMSE do so from early in training and continue to track dMMSE as the task distribution changes.

\begin{figure*}[t]
  \centering
  \includegraphics[width=\textwidth]{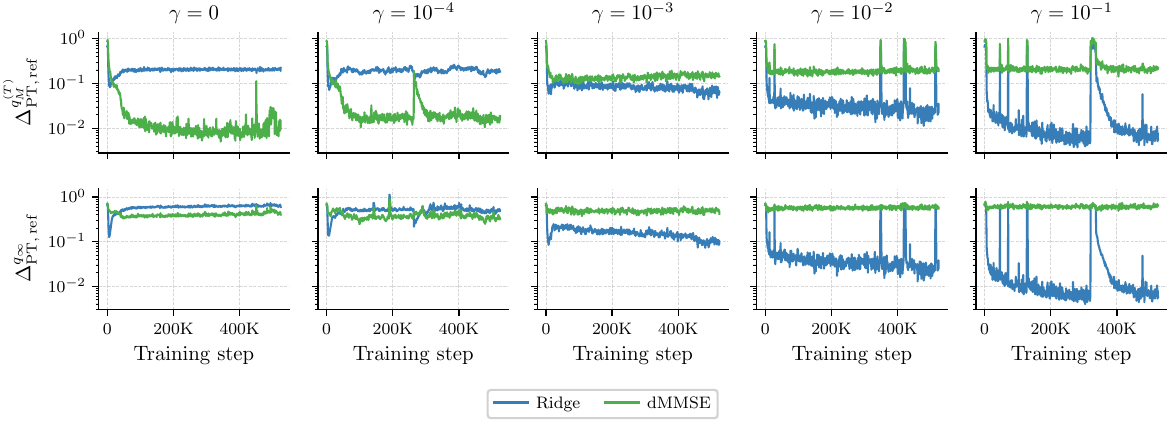}
  \caption{\textbf{Predictions of transformers below the non-stationary task diversity threshold track dMMSE throughout training.} We show the mean squared distance $\Delta_\text{PT,dMMSE}$ and $\Delta_\text{PT,Ridge}$ throughout training.
  We use a fixed task diversity $M = 32$ and MALA step sizes
    $\gamma \in \{0, 10^{-4}, 10^{-3}, 10^{-2}, 10^{-1}\}$.
  We evaluate on in-distribution sequences from
  $q_M^{(\tau)}$ (\textit{top row}) and on
  out-of-distribution sequences from
  $q_\infty$ (\textit{bottom row}).
  The results are from a single training seed.
  The first two columns correspond to transformers below the task diversity threshold for their respective step sizes, and we see that in terms of in-distribution mean squared distance their predictions track the moving dMMSE reference predictor throughout most of training.
  See \cref{fig:time-delta-metrics-dirichlet} for the resampling non-stationarity analogue.
}
  \label{fig:time-delta-metrics}
\end{figure*}

\subsection{Perspectives on transformer behaviour}
\label{sec:pursuit:perspectives}

We analyse our transformers' behavioural evolution by measuring distances from the reference predictors in two spaces.
\begin{enumerate}
    \item
        \textbf{Prediction space \textnormal{(\cref{sec:pursuit:deltas})}:}
        We track the mean square difference $\Delta^q_{\text{PT}, \texttt{ref}}$ between the predictions of each transformer checkpoint and each reference predictor~(\ref{eq:deltas}).
    \item 
        \textbf{Implicit prior space \textnormal{(\cref{sec:pursuit:predictive_resampling,sec:pursuit:results})}:}
        We estimate the transformers' implicit prior over latent regression tasks using predictive Monte Carlo \citep{Fortini+Petrone2023,Fortini+Petrone2025,EffiezalAswadi+2026}. We track energy distance \citep{Szekely1989,Szekely+Rizzo2013} between these distributions and those of the reference predictors.
\end{enumerate}
Predictive Monte Carlo requires the model to output a full predictive distribution rather than a traditional point prediction for $\hat y_k$. Therefore, for the results in this section, we train with a variant architecture replacing the transformer's point-prediction head with an affine transform that outputs the parameters of a mixture of Gaussians. At each query position, the model outputs mixture weights $\{\pi_g\}_{g=1}^G$, means $\{\mu_g\}_{g=1}^G$, and variances $\{\sigma_g^2\}_{g=1}^G$, defining the predictive distribution
\begin{equation}
    f(y \mid S_{\leq k}) = \sum_{g=1}^{G} \pi_g \, \mathcal{N}(y; \mu_g, \sigma_g^2).
\end{equation}
We fix the number of mixture components $G=4$ across all models. We train by minimising the negative log-likelihood  \citep{Bishop1994}, using a $\operatorname{logsumexp}$ trick for numerical stability (see \cref{app:gmm-head} for details).
We also train on longer sequences ($K = 64$ in-context examples) to provide sufficient rollout length for predictive Monte Carlo.

\subsection{Prediction trajectories}
\label{sec:pursuit:deltas}

In this section, we use the random walk non-stationarity setting of \cref{sec:experiments:mala} with MALA step size $\gamma \in \{0,10^{-4},10^{-3},10^{-2},10^{-1}\}$.
We also focus on a fixed task diversity of $M=32$. This task diversity lies below the task diversity threshold in the stationary setting, but above the threshold for sufficiently high $\gamma$. It is therefore interesting to see how different values of $\gamma$ affect the model's behaviour throughout training.
See \cref{app:grid-plots} for analysis of resampling non-stationarity and all task diversities.

\cref{fig:time-delta-metrics} shows the evolution of a transformer's behaviour throughout training. At low step sizes ($\gamma \in \{ 0,10^{-4}\}$) the transformer's predictions track those of the dMMSE reference predictor. At larger step sizes ($\gamma \in \{ 10^{-3}, 10^{-2},10^{-1}\}$), we see the model transitioning to predicting more like the ridge predictor.

For most of training the mean squared differences are approximately stable for each model. Near the beginning of training we see some models briefly trend towards ridge predictions before pivoting back towards dMMSE.
This parallels the transient ridge phenomenon \citep[see also \citealp{Singh+2024}]{Panwar+2024,Carroll+2025}.

\begin{figure*}[t]
  \centering
  \includegraphics[width=\textwidth]{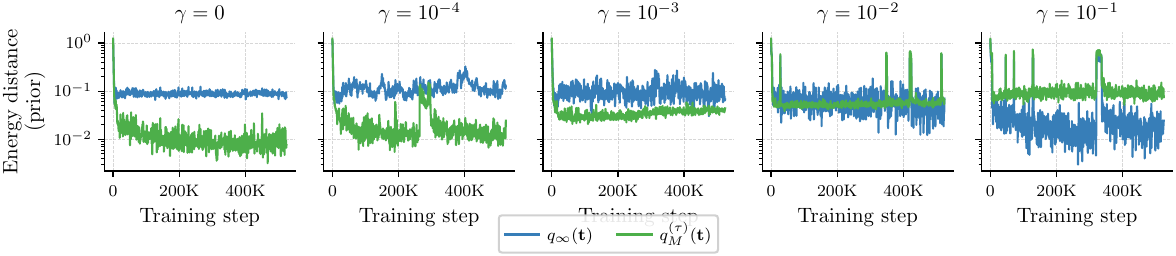}
  \caption{\textbf{Predictive Monte Carlo reveals that below the non-stationary task diversity threshold, the transformer's implicit task distribution tracks the changing task distribution.}
  We use predictive Monte Carlo to extract the transformer's implicit prior over task vectors throughout training, and compare to the finite task distribution $q^{(\tau)}_M(\mathbf{t})$ and infinite task diversity $q_\infty(\mathbf{t}) = \mathcal N(0, I_D)$ priors via energy distance.
  As in \cref{fig:time-delta-metrics}, we fix $M = 32$ and vary the MALA step size $\gamma \in \{0, 10^{-4}, 10^{-3}, 10^{-2}, 10^{-1}\}$.
  The results are from a single training seed.
  The first two columns correspond to transformers below the task diversity threshold for their respective step sizes, and we see that the revealed priors over tasks closely track the changing task distribution for most of training.
  See \cref{fig:time-energy-distance-dirichlet} for the resampling non-stationarity analogue.
  }
  \label{fig:time-energy-distance}
\end{figure*}

\subsection{Predictive Monte Carlo}
\label{sec:pursuit:predictive_resampling}

Predictive Monte Carlo allows us to sample from the implicit prior and posterior of a Bayes-filtered transformer, as shown in \citet{EffiezalAswadi+2026}. We briefly review this methodology, following \citet{Fortini+Petrone2023,Fortini+Petrone2025}. In a supervised setting, a \emph{predictive rule} is a sequence of conditional distributions 
    $f_k(\cdot) = p(y_{k+1} \in \cdot \mid S_{\leq k + 1})$.
Under (sufficient but not necessary) regularity conditions 
    \citep{Fortini+Petrone2023,Fong+2023,Battiston+2025},
the predictive distributions converge almost surely along a rollout:
    $f_L \Rightarrow \tilde{F}$ as $L \to \infty$, 
where $\tilde{F}$ is a directing measure over observations. 

Our transformer provides a predictive rule for labels $y_k$ conditioned on $S_{\le k}$, but not the input distribution. We therefore construct each rollout by alternating between sampling inputs from the known input distribution and sampling labels from the transformer: for $k = 1$, \dots, $L$, we draw $x_k \sim \mathcal{N}(0, I_D)$ and then draw $y_k \sim f(\cdot \mid S_{\leq k})$ from the transformer's predictive distribution.

To recover the transformer's implicit prior over latent tasks, we follow the approach developed by \citet{EffiezalAswadi+2026}. For large rollout length $L$, the generated sequence from a predictive rule $(x_1, y_1), \dots, (x_L, y_L)$ approximates an i.i.d.\ sample from $\tilde{F}$. In the linear regression case, $\tilde{F}$ is parametrised by a task vector $\mathbf{t}$. Since OLS is the maximum likelihood estimator, fitting OLS to the generated sequence gives an estimate of the task vector $\mathbf{t}$ that parametrises
$\tilde{F}$. Starting from an empty context produces samples from the
implicit prior over $\mathbf{t}$.
Each rollout converges to a different realisation of $\tilde{F}$, so repeating $N$ times produces a Monte Carlo sample $\{\hat{\mathbf{t}}^{(i)}\}_{i=1}^N$ from the transformer's implicit prior over task vectors.

\subsection{Implicit prior trajectories}
\label{sec:pursuit:results}
\label{sec:pursuit:visualisation}

We study the same random-walk non-stationarity setting for task diversity $M=32$ (see \cref{app:grid-plots} for resampling and all task diversities).
For predictive Monte Carlo, we sample from the implicit prior of the transformer (rolling out without a prompt). We note that we do not check the conditions sufficient for convergence. We compute energy distance using Monte Carlo samples from the reference priors.

\Cref{fig:time-energy-distance} shows that at low step sizes ($\gamma \in \{0,  10^{-4}\}$), below the respective task diversity thresholds, the transformers' revealed priors closely track the task distribution in terms of energy distance.
At intermediate step sizes ($\gamma \in \{10^{-3}, 10^{-2}\}$) the revealed prior is somewhere between the instantaneous task distribution $q^{(\tau)}_{32}(\mathbf t) = \text{Uniform}(\mathcal T^{(\tau)}_{32})$ and the time-average task distribution $q_\infty(\mathbf t) = \mathcal N(0,I_D)$.
At step size $\gamma = 10^{-1}$ the model's implicit prior matches closely the time-average task distribution.
These results complement \cref{fig:time-delta-metrics} by showing that the same shift occurs in latent task distribution space, not only in prediction space.

\begin{figure*}[t]
  \centering
  \includegraphics[width=\textwidth]{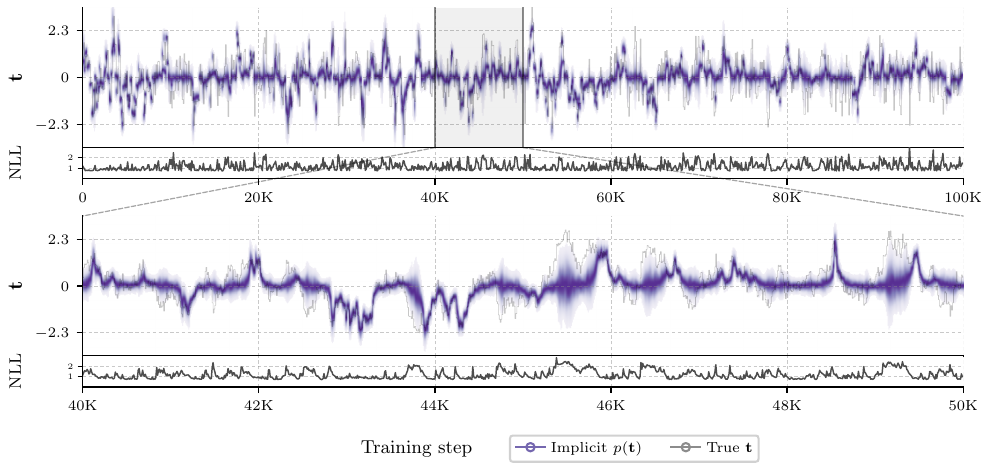}
  \caption{\textbf{Implicit prior over a 1D task vector against the true task during training.} We use predictive Monte Carlo to extract the transformer's implicit prior over the task vector $p(\mathbf t)$ (purple) and compare it to the true task $\mathbf t$ (black), for a one dimensional MALA setting with task dimension $D = 1$, task diversity $M = 1$, and $\gamma = 10^{-2}$. The \textit{top panel} shows $0$ to $100$K training steps, and the \textit{bottom panel} zooms into the $40$K to $50$K interval. Below each plot we include the (per-token mean) negative log-likelihood on the training batch.}
  \label{fig:prior-over-time-callout}
\end{figure*}

Computing energy distance from our reference distributions summarises the implicit prior with a pair of scalars. For a one-dimensional task distribution ($D=1$), we can visualise the entire implicit prior using a histogram. \Cref{fig:prior-over-time-callout} displays the evolution of this implicit prior histogram over training time for a transformer trained against a single varying task (task diversity $M=1$, MALA step size $\gamma=10^{-2}$). We train for only 100K steps, sampling more densely in the 40K--50K interval to observe the distribution's dynamics more clearly. For this configuration, the prior sometimes tracks the changing task, whereas other times it remains concentrated at the origin while the task varies, suggesting we are near a non-stationary task diversity threshold. See \cref{app:grid-plots} for all step sizes and resampling non-stationarity.

\section{Explanations}
\label{sec:discussion}

In this section, we discuss several possible qualitative explanations for the phenomena studied in \cref{sec:experiments,sec:experiments:pursuit}.
We first dismiss two hypotheses that suggest the transformer selects its algorithm based on all tasks seen during training (\S\ref{sec:discussion:bayes}, \S\ref{sec:discussion:all}).
We then offer two more nuanced hypotheses, based on diversity amongst tasks seen in recent steps (\S\ref{sec:discussion:window}) and a bias towards inherently stable solutions (\S\ref{sec:discussion:stability}).
Our observations fail to distinguish between these hypotheses, but we propose considerations for future experiments (\S\ref{sec:discussion:future}).

\subsection{Online Bayesian model selection}
\label{sec:discussion:bayes}

Bayesian inference is an idealised model of learning and therefore a natural candidate model of deep learning phenomena. In the setting of in-context linear regression specifically, \citet{Carroll+2025} and \citet{Wurgaft+2025} appeal to Bayesian model selection to explain the dynamic trade-off we observe between dMMSE and ridge solutions.

It is straightforward to model online learning in a Bayesian framework.
Suppose we draw data $S_1, \ldots, S_T$ independently from a time-varying distribution $S_\tau \sim q^{(\tau)}$. Given a prior $\pi_0(\theta)$ over some model class, derive a sequence of posteriors using the incremental form of Bayes' rule,
    $\pi_\tau(\theta) = \mathbb{P}(S_\tau | \theta) \pi_{\tau-1}(\theta) / \mathbb{P}(S_\tau | S_1, \ldots, S_{\tau-1})$.
Observe, however: this online formulation is equivalent to full-batch Bayesian inference,
    $\pi_T(\theta) = \prod_{\tau=1}^T\mathbb{P}(S_\tau | \theta) \pi_0(\theta) / \prod_{\tau=1}^T\mathbb{P}(S_\tau)$,
and therefore invariant to data order.
Thus, this model cannot predict a distinction between stationary and non-stationary learning if the long-term time average of the non-stationary distributions matches the stationary distribution.

Both of our non-stationary learning settings have a long-term time average task distribution of $\mathcal N(0, I_D)$. Given enough training to realise the long-term time average, this should cause the transformers to always learn the ridge solution \citep{Raventos+2023}. Since we observe our transformers tracking the dMMSE solution for hundreds of thousands of steps, we rule out this model.

\subsection{Total task diversity}
\label{sec:discussion:all}

Under the non-stationary condition, our transformers see many more tasks over training than in the stationary setting with the same finite task diversity.
\citet{Raventos+2023} already showed that transformers that see a larger number of tasks during training are more likely to prefer the ridge solution.
Maybe in the non-stationary setting the choice of solution is determined not by the \emph{instantaneous} task diversity $M$, but by the \emph{total} number of tasks seen throughout training---call the latter the \emph{total task diversity}.

For resampling non-stationarity (\S\ref{sec:experiments:dirichlet}) the number of distinct tasks seen during training is almost surely $R \times M$.
For random walk non-stationarity (\S\ref{sec:experiments:mala}), the appropriate effective task diversity is less clear: with small step sizes acceptance probabilities are usually very high but the differences between tasks is extremely small; with sufficiently large step sizes the distinctions are more significant but acceptance probabilities drop. 
We could use the integrated autocorrelation time of the random walk as a proxy for the number of steps required to sample a distinct task. Since we are sampling from an $8$-dimensional isotropic Gaussian, this quantity should be a small constant.

However, this model predicts much more aggressive lowering of the task diversity threshold than we observe. The total task diversity for many of our training runs exceeds the stationary task diversity threshold, yet we see many examples of these transformers converging to the dMMSE solution. Moreover, these transformers pursue the instantaneous dMMSE throughout training. Therefore, we dismiss the total task diversity hypothesis.

\subsection{Effective task diversity within recent memory}
\label{sec:discussion:window}

In light of the literature on catastrophic forgetting \citep{McCloskey+Cohen1989,Ratcliff1990,French1999,Goodfellow+2014}, it is not surprising that our transformers are insensitive to tasks encountered early in training.

We can refine the total task diversity hypothesis by supposing the transformer's selection \emph{is} made based on the stationary task diversity threshold, but against an \emph{effective task diversity} given by the number of tasks seen recently (with the exact meaning of ``recently'' to be defined).
For example, one could posit a fixed number of recent training steps and take the task diversity of the average distribution. Or, one could count all tasks seen but decay their contribution exponentially as they are replaced by new tasks.

For the right definition of effective task diversity, it should be possible to overcome the aggressive predictions of the total task diversity hypothesis.

\subsection{Optimisation tends towards more stable models}
\label{sec:discussion:stability}

Finally, we offer an alternative hypothesis---in the presence of non-stationarity, deep learning optimisation trajectories tend towards models that generalise across time and away from models that must be continually adapted in the face of data distribution changes.

The mechanism for this hypothetical tendency would be that internal model structure that contributes towards stable predictions will be continually reinforced during gradient-based training, while structure supporting instantaneous specialisation has to be continually re-learned.

This hypothesis correctly predicts the qualitative shift in task diversity we see: the ridge solution is invariant to the changes in the task distribution we consider, whereas the dMMSE solution repeatedly changes along with the task distribution, forcing a transformer that prefers dMMSE to continually pursue a changing set of tasks.

\subsection{Towards distinguishing the latter explanations}
\label{sec:discussion:future}

The qualitative observation that increasing non-stationarity decreases the task diversity threshold is insufficient to distinguish between the effective task diversity hypothesis and the stability hypothesis. This is a limitation of our experimental design, since increasing non-stationarity coincides with increasing effective task diversity in our set-up.

Future work should design further experiments to isolate the contributions of these (or other) explanations to the observed phenomenon, to the extent possible. We offer some preliminary thoughts in this direction.
\begin{enumerate}
    \item
        The challenge is to design an experiment where a change in stability comes apart from a change in effective task diversity. It may be possible to achieve this by considering a data distribution constructed by cycling through a fixed pool of tasks. Cycling tasks makes memorising the current set of tasks unstable, but effective task diversity is capped at the size of the task pool.
    \item
        Alternatively, it may be supposed that, regardless of the precise formalisation of the effective task diversity, the degree of sensitivity of the transformer to recent tasks is dependent on the configuration of the architecture and optimiser (e.g., learning rate) and independent of the choice of data. It may be possible to fit the parameters of a ``sensitivity window'' at one level of non-stationarity and then see whether these parameters continue to be predictive at another---any discrepancy would point to a contribution from another explanation.
\end{enumerate}

We note that \emph{both} the effective task diversity hypothesis and the stability hypothesis may have roles to play in a full explanation of the phenomenon we have studied. Moreover, they may overlap in that they turn out to offer different perspectives on similar underlying learning dynamics.

\section{Conclusion}

We have demonstrated that increasing non-stationarity decreases the task diversity threshold for in-context linear regression transformers.
Below this reduced task diversity threshold, transformers continuously memorise an ever-changing set of latent regression tasks. Above the reduced task diversity threshold, transformers instead learn the stable ridge solution that generalises across time.

This \emph{temporal task diversity} phenomenon provides a new window into the increasingly important topic of the inductive biases of modern deep learning algorithms, in the important setting of non-stationary deep learning.
We invite future work that seeks to more precisely characterise and explain this phenomenon.

\section*{Impact statement}

This paper presents work whose goal is to advance the science of deep learning by uncovering a novel phenomenon that arises under non-stationarity. While further work is needed to reveal the causes of this phenomenon, we hope that eventually a greater understanding of the nature of the influences of changing data sets on learning in practical settings can contribute to more robust alignment methodologies, along the lines of \citet{PepinLehalleur+2025}.

\section*{Reproducibility statement}

Experiment details are summarised in \cref{app:experiment-details}. For code required to replicate data generation, transformer training, and predictive Monte Carlo methodologies please see \href{https://github.com/matomatical/temporal-task-diversity}{github.com/\-matomatical/\-temporal-\-task-\-diversity}.

\section*{Acknowledgements}

For helpful conversations, we thank 
  Chris Elliott,
  Lorenz Hufe,
  Edmund Lau,
  Daniel Murfet,
  Susan Wei,
and 
  Joan Velja.
Research supported with Cloud TPUs from Google's TPU Research Cloud (TRC).


\balance
\bibliographystyle{icml2026}
\bibliography{references}

\clearpage
\appendix
\crefalias{section}{appendix}
\onecolumn

\begin{table*}[!b]
\centering
\caption{\textbf{Summary of hyper-parameters.} Where a single hyper-parameter is shared across all experiments, we omit repetitions (---).}
\label{tab:hyperparams}
\begin{tabular}{@{}llccc@{}}
\toprule
\textbf{Category} & \textbf{Hyper-parameter} & \textbf{\S\ref{sec:experiments:mala}} & \textbf{\S\ref{sec:experiments:dirichlet}} & \textbf{\S\ref{sec:experiments:pursuit}} \\
\midrule
Data      & Task dimension ($D$)              & 8       & ---     & --- \\
Data      & In-context examples ($K$)         & 16      & 16      & 64 \\
Data      & Observation noise variance ($\sigma^2$) & 0.25 & --- & --- \\
\midrule
Model     & Number of layers                  & 8       & ---     & --- \\
Model     & Number of attention heads per layer & 2     & ---     & --- \\
Model     & Embedding dimension               & 128     & ---     & --- \\
Model     & MLP hidden dimension              & 128     & ---     & --- \\
Model     & Layer normalisation               & Pre     & ---     & --- \\
Model     & Prediction head type              & Point   & Point   & Mixture density \\
Model     & Mixture components ($G$)          & & & 4 \\
\midrule
Optimiser & Type of optimiser                 & Adam    & ---     & --- \\
Optimiser & Moment estimate decay rates ($\beta_1$,$\beta_2$) & $(0.9, 0.999)$ & --- & --- \\
Optimiser & Weight decay                      & No      & ---     & --- \\
\midrule
Learning rate schedule & Shape                & Increase then constant & --- & --- \\
Learning rate schedule & Peak learning rate ($\eta$)       & $3 \times 10^{-3}$ & --- & --- \\
Learning rate schedule & Warm-up strategy      & Linear  & ---     & --- \\
Learning rate schedule & Warm-up fraction      & $10\%$  & ---     & --- \\
\midrule
Experiments & Batch size ($B$)                & 256     & ---     & --- \\
Experiments & Training steps ($T$)            & 524,000 & ---     & --- \\
Experiments & Number of training seeds        & 7       & 7       & 1 \\
\midrule
Predictive Monte Carlo & Number of rollouts    & & & 5,000 \\
Predictive Monte Carlo & Rollout length        & & & 64 \\
\bottomrule
\end{tabular}
\end{table*}

\section{Experiment details}
\label{app:experiment-details}

\paragraph{Architecture.}
We use a decoder-only transformer with pre-layer normalisation, learnable positional embeddings, and causal masking \citep[see, e.g.,][]{Elhage+2021,Phuong+Hutter2022}.
Specifically, we use 8 layers, two attention heads per layer, and an embedding and MLP dimension of 128. A final layer normalisation precedes either a learned affine output head that maps to a scalar prediction in \cref{sec:experiments:mala,sec:experiments:dirichlet}, or to a mixture density head which maps to the means, variances, and mixture weights of a $G$-component Gaussian mixture in \cref{sec:experiments:pursuit}.

\paragraph{Input tokenisation.}
Following \citet{Raventos+2023}, we encode each sequence $S = (x_1, y_1, \ldots, x_K, y_K)$ as a sequence of $2K$ input tokens in $\mathbb R^{D+1}$:
\begin{equation*}
\left(
    \begin{pmatrix} 0\\ x_1 \end{pmatrix}\!,
    \begin{pmatrix} y_1 \\ 0 \end{pmatrix}\!,
    \begin{pmatrix} 0 \\ x_2 \end{pmatrix}\!,
    \begin{pmatrix} y_2 \\ 0 \end{pmatrix}\!,
    \ldots,
    \begin{pmatrix} 0 \\ x_{K} \end{pmatrix}\!,
    \begin{pmatrix} y_K \\ 0 \end{pmatrix}
\right).
\end{equation*}

\paragraph{Output tokenisation.}
We output token sequences of length $2K$, but discard every second token so as to only consider one prediction per regression example.
Each output token is either a scalar prediction $\tilde y_k$ (for the usual point prediction setting) or $3G$ scalars (for the mixture density setting, see \cref{sec:pursuit:perspectives}).

\paragraph{Training.}
We train each transformer for $T = 524\text{K}$ steps using a batch size of $B = 256$. We optimize with Adam \citep{Kingma+Ba2015} without weight decay. We use a learning rate schedule that increases linearly from $0$ to its maximum value $\eta$ over a fraction of training steps and then remains constant. For \cref{sec:experiments:mala,sec:experiments:dirichlet}, the loss is mean squared error over all $K$ in-context examples. For \cref{sec:experiments:pursuit}, the loss is negative log-likelihood under the mixture density head.

\paragraph{Compute.}
We use a bespoke transformer implementation modelled after NanoGPT \citep{Karpathy2022} and using JAX \citep{Bradbury+2018}.
Training one model (524K batches of 256 sequences) took approximately 45~minutes on a single dual-core TPU~v4 device.
Predictive Monte Carlo for one checkpoint (5K rollouts of length 64) took around 25~TPU-device-seconds. Each column in \cref{fig:time-energy-distance} involves 656 checkpoints, thus taking approximately 5~hours (plus transformer training time).
TPUs were provided by Google's TPU Research Cloud.

\clearpage

\section{Derivation of mixture density head}
\label{app:gmm-head}

We replace the point prediction head with a learned affine map to $\mathbb{R}^{3G}$, parametrising a Gaussian mixture model, as in \citet{Bishop1994}. Following the tokenisation scheme described in \cref{app:experiment-details}, we keep only the predictions at query positions (every second token). Each prediction vector is partitioned into $G$ separate logits $z^\pi_g$, means $\mu_g$, and raw parameters $z^\sigma_g$. We map $z^\sigma_g$ to a positive real number representing the variance using the softplus function, $\sigma_g^2 = \operatorname{softplus}(z^\sigma_g)$. The logits could be converted to probabilities $\pi_g$ using $\operatorname{softmax}$ though in practice we never use this as we work in log-space in our implementations.

The loss function is the negative log-likelihood of the observed $y$ under the predicted mixture:
\[
    -\log p(y \mid \pi, \mu, \sigma^2) = -\log \sum_{g=1}^{G} \pi_g \, \mathcal{N}(y;\, \mu_g,\, \sigma_g^2).
\]
For numerical stability, we rewrite each term inside the sum as
\[
    \pi_g \, \mathcal{N}(y;\, \mu_g,\, \sigma_g^2) = \exp\!\left(\log \pi_g - \mathrm{NLL}_g\right),
\]
where
\[
    \mathrm{NLL}_g = \frac{1}{2}\left(\log(2\pi\sigma_g^2) + \frac{(y - \mu_g)^2}{\sigma_g^2}\right)
\]
is the (Gaussian) negative log-likelihood for component $g$. This then gives
\[
    -\log p(y \mid \pi, \mu, \sigma^2) = -\operatorname{logsumexp}_g\!\left(\log \pi_g - \mathrm{NLL}_g\right),
\]
which avoids evaluating small exponentials directly.

\clearpage

\section{Additional plots}
\label{app:grid-plots}

\begin{figure*}[!b]
  \centering
  \includegraphics[width=\textwidth]{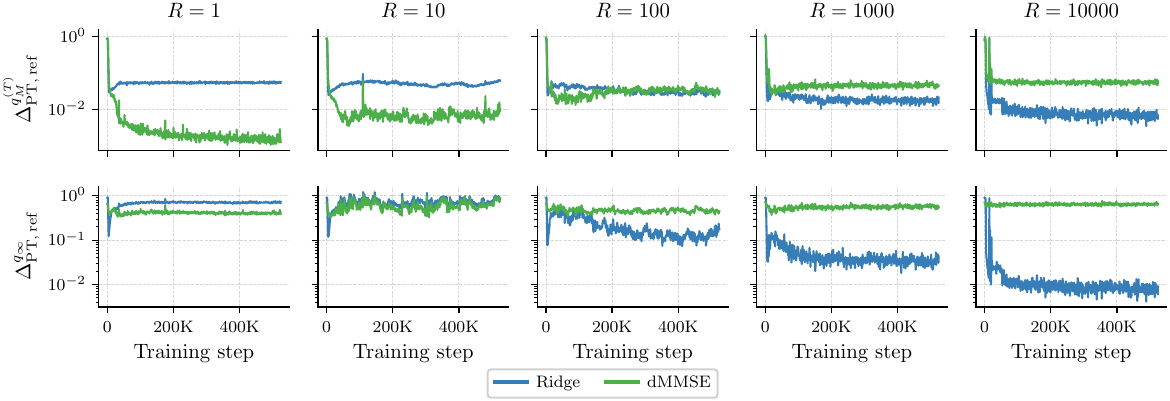}
  \caption{
  Resampling non-stationarity analogue of \cref{fig:time-delta-metrics}, for task diversity $M = 32$ and sample numbers $R \in \{1, 10, 100, 1000, 10000\}$. The first two columns correspond to transformers below the task diversity threshold for their respective resampling number $R$, and we see that in terms of in-distribution mean squared distance their predictions track the moving dMMSE reference predictor throughout most of training.}
  \label{fig:time-delta-metrics-dirichlet}
\end{figure*}

\begin{figure*}[!b]
  \centering
  \vspace{0.25\baselineskip}
  \includegraphics[width=\textwidth]{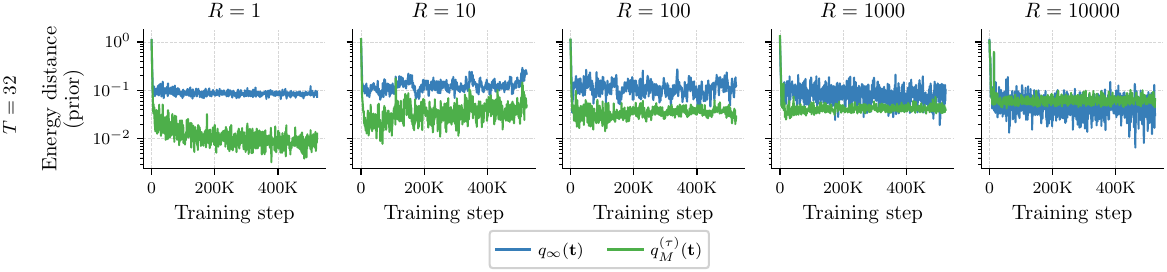}
  \caption{%
  Resampling non-stationarity analogue of \cref{fig:time-energy-distance}, for task diversity $M = 32$ and sample numbers $R \in \{1, 10, 100, 1000, 10000\}$. The first two columns correspond to transformers before the task diversity threshold for their respective resampling number $R$, and we see that the revealed priors over tasks closely track the changing task distribution for most of training.}
  \label{fig:time-energy-distance-dirichlet}
\end{figure*}

\Cref{fig:time-delta-metrics-dirichlet,fig:time-energy-distance-dirichlet} show the resampling non-stationarity analogues of \cref{fig:time-delta-metrics,fig:time-energy-distance} from \cref{sec:experiments:pursuit}, exhibiting the same qualitative behaviour.

\begin{figure*}[p]
  \centering
  \includegraphics[width=\textwidth,height=0.9\textheight,keepaspectratio]{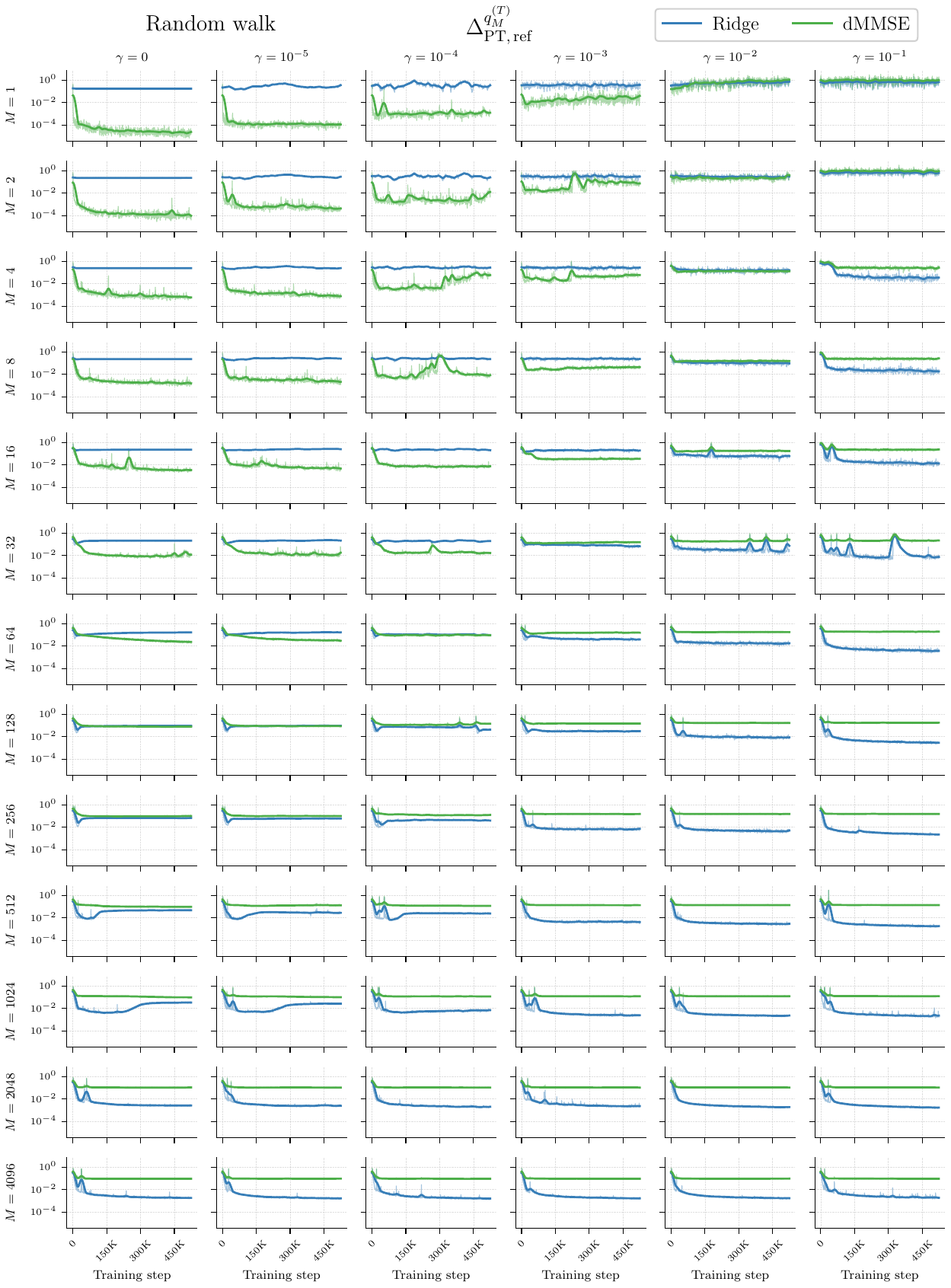}
  \caption{\textbf{In-distribution mean squared prediction differences throughout training under random walk non-stationarity.} We show $\Delta_\text{PT,dMMSE}$ and $\Delta_\text{PT,Ridge}$ on in-distribution sequences from $q_M^{(\tau)}$ for each combination of task diversity $M$ and MALA step size $\gamma$.}
  \label{fig:mala-grid-delta-id}
\end{figure*}

\begin{figure*}[p]
  \centering
  \includegraphics[width=\textwidth,height=0.9\textheight,keepaspectratio]{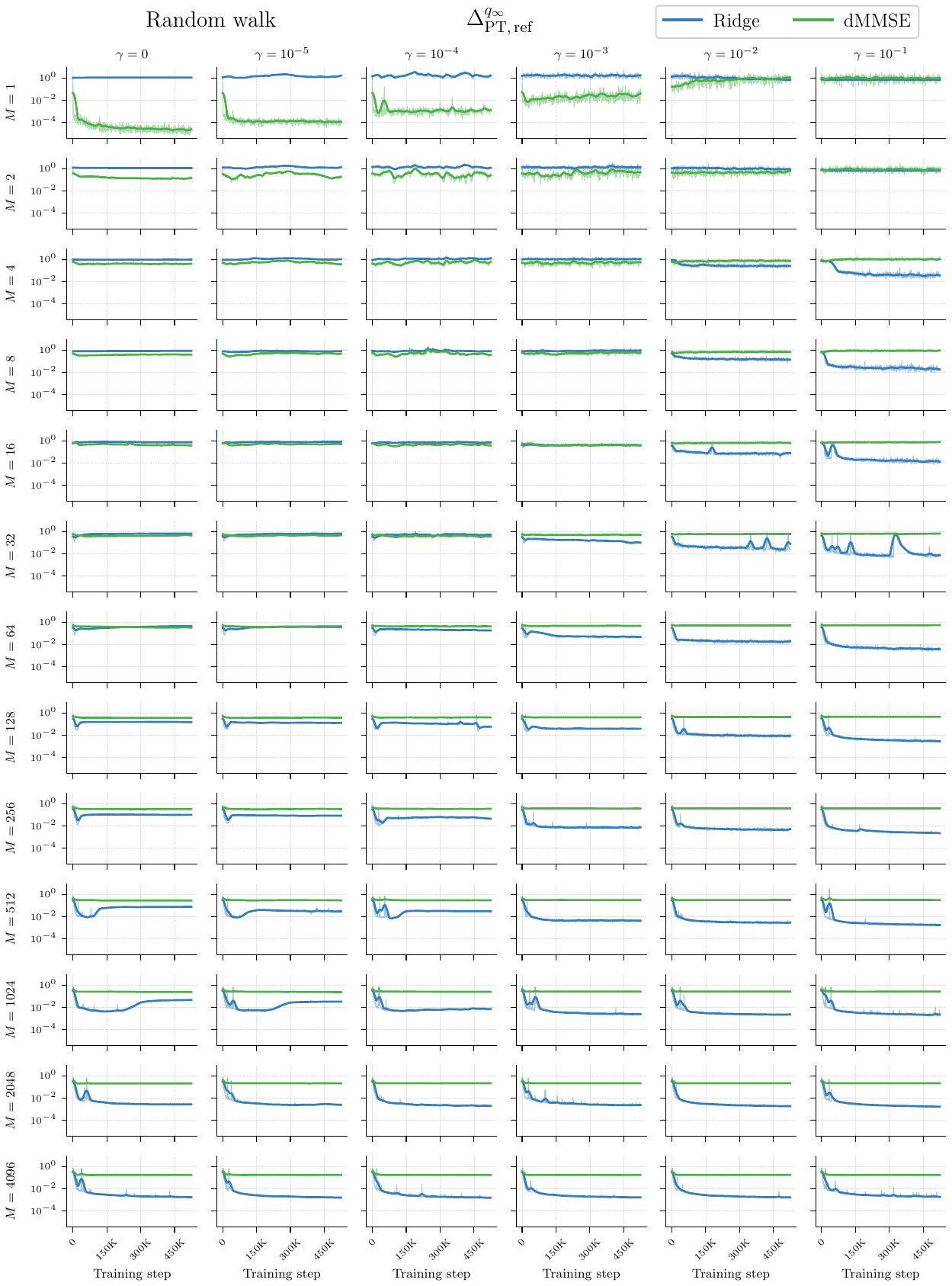}
  \caption{\textbf{Out-of-distribution mean squared prediction differences throughout training under random walk non-stationarity.} We show $\Delta_\text{PT,dMMSE}$ and $\Delta_\text{PT,Ridge}$ on out-of-distribution sequences from $q_\infty$ for each combination of task diversity $M$ and MALA step size $\gamma$.}
  \label{fig:mala-grid-delta-ood}
\end{figure*}

\begin{figure*}[p]
  \centering
  \includegraphics[width=\textwidth,height=0.9\textheight,keepaspectratio]{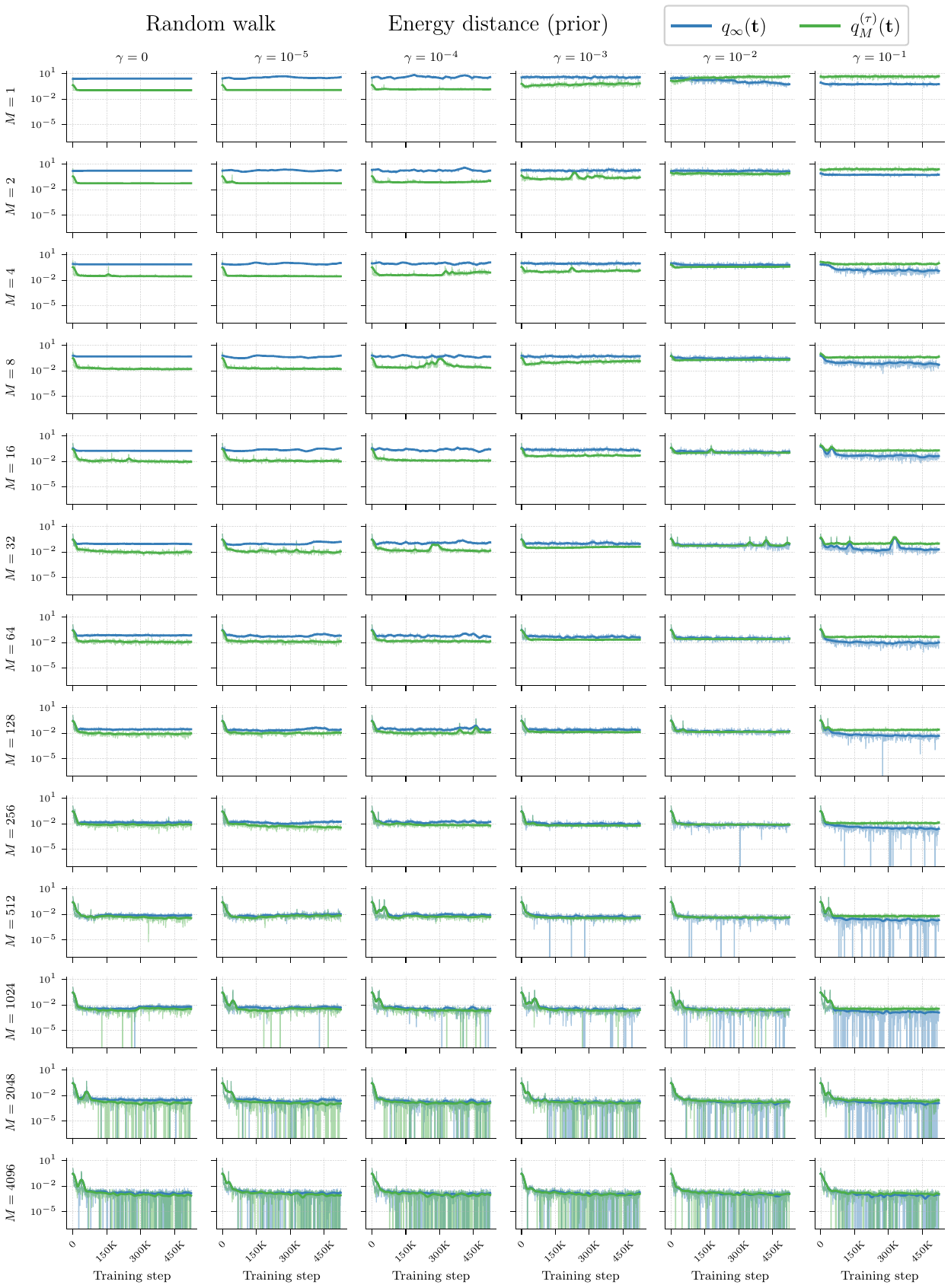}
  \caption{\textbf{Energy distance between the transformer's implicit prior and the $\text{Uniform}(\mathcal T^{(\tau)}_M)$ and $\mathcal N(0, I_D)$ priors throughout training under random walk non-stationarity.} For each combination of task diversity $M$ and MALA step size $\gamma$, we use predictive Monte Carlo to extract the transformer's implicit prior over task vectors throughout training, and compare to the baseline priors via energy distance.}
  \label{fig:mala-grid-energy}
\end{figure*}

\begin{figure*}[p]
  \centering
  \includegraphics[width=\textwidth,height=0.9\textheight,keepaspectratio]{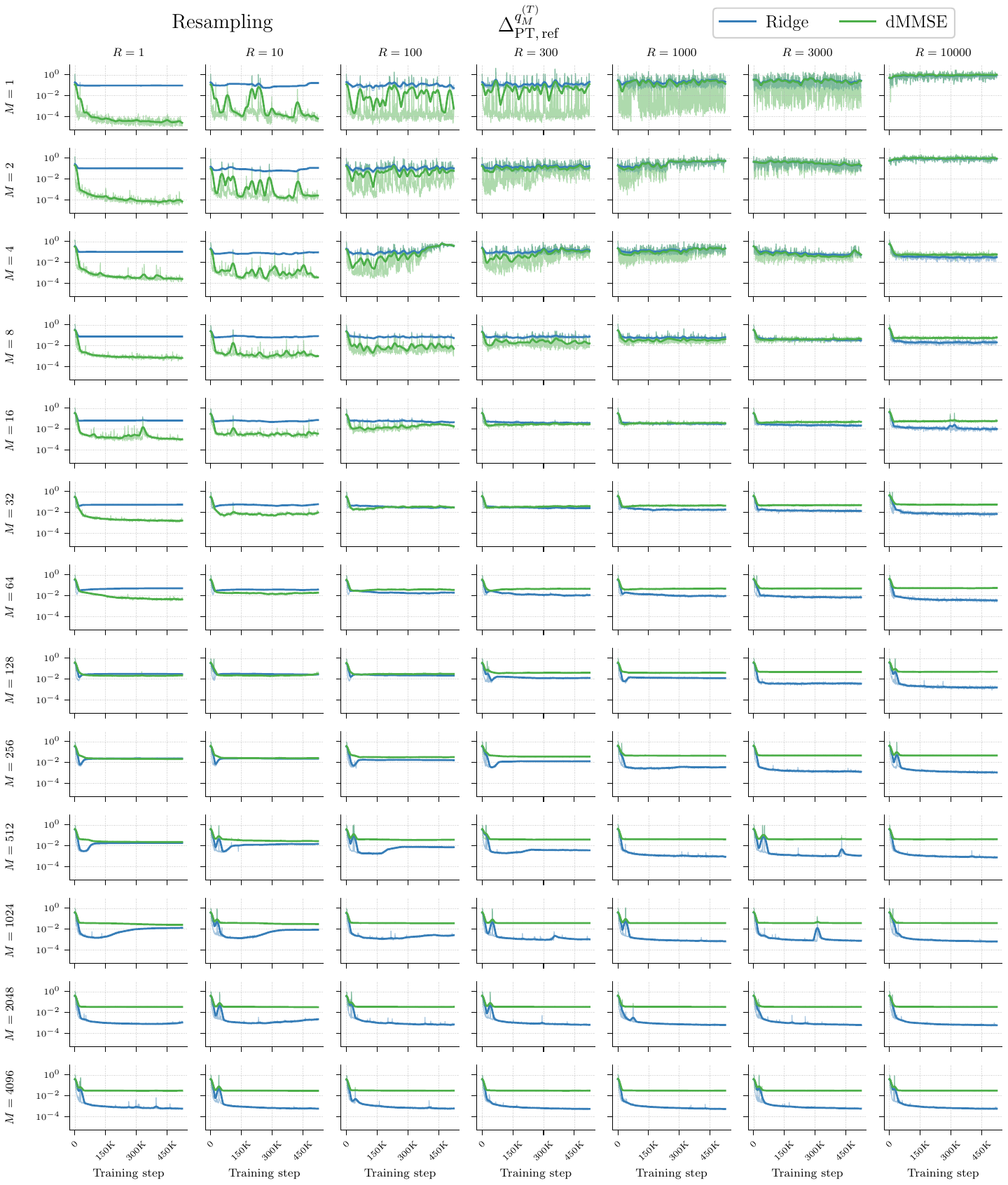}
  \caption{\textbf{In-distribution mean squared prediction differences throughout training under resampling non-stationarity.} We show $\Delta_\text{PT,dMMSE}$ and $\Delta_\text{PT,Ridge}$ on in-distribution sequences from $q_M^{(\tau)}$ for each combination of task diversity $M$ and sample number $R$.}
  \label{fig:dirichlet-grid-delta-id}
\end{figure*}

\begin{figure*}[p]
  \centering
  \includegraphics[width=\textwidth,height=0.9\textheight,keepaspectratio]{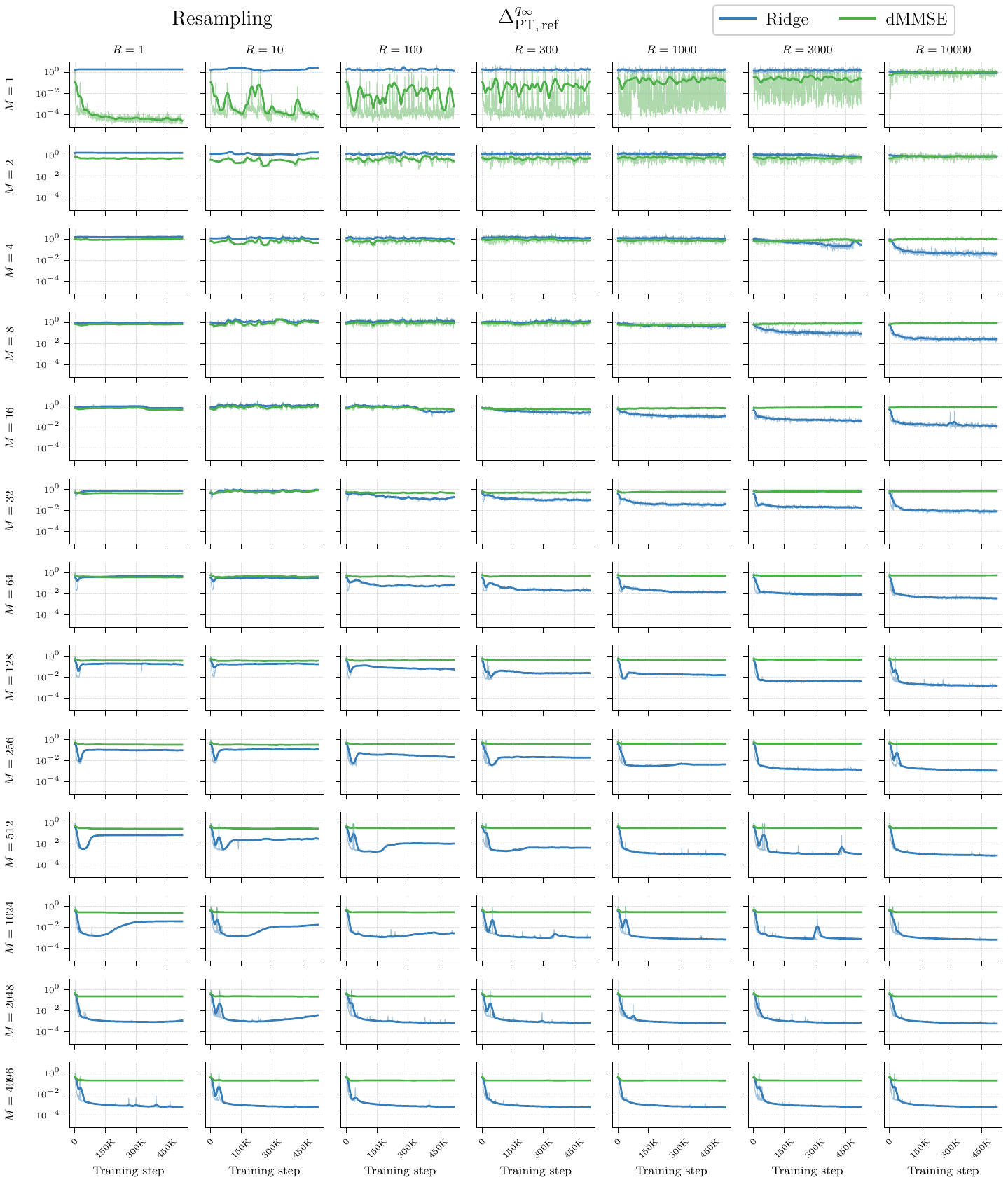}
  \caption{\textbf{Out-of-distribution mean squared prediction differences throughout training under resampling non-stationarity.} We show $\Delta_\text{PT,dMMSE}$ and $\Delta_\text{PT,Ridge}$ on out-of-distribution sequences from $q_\infty$ for each combination of task diversity $M$ and sample number $R$.}
  \label{fig:dirichlet-grid-delta-ood}
\end{figure*}

\begin{figure*}[p]
  \centering
  \includegraphics[width=\textwidth,height=0.9\textheight,keepaspectratio]{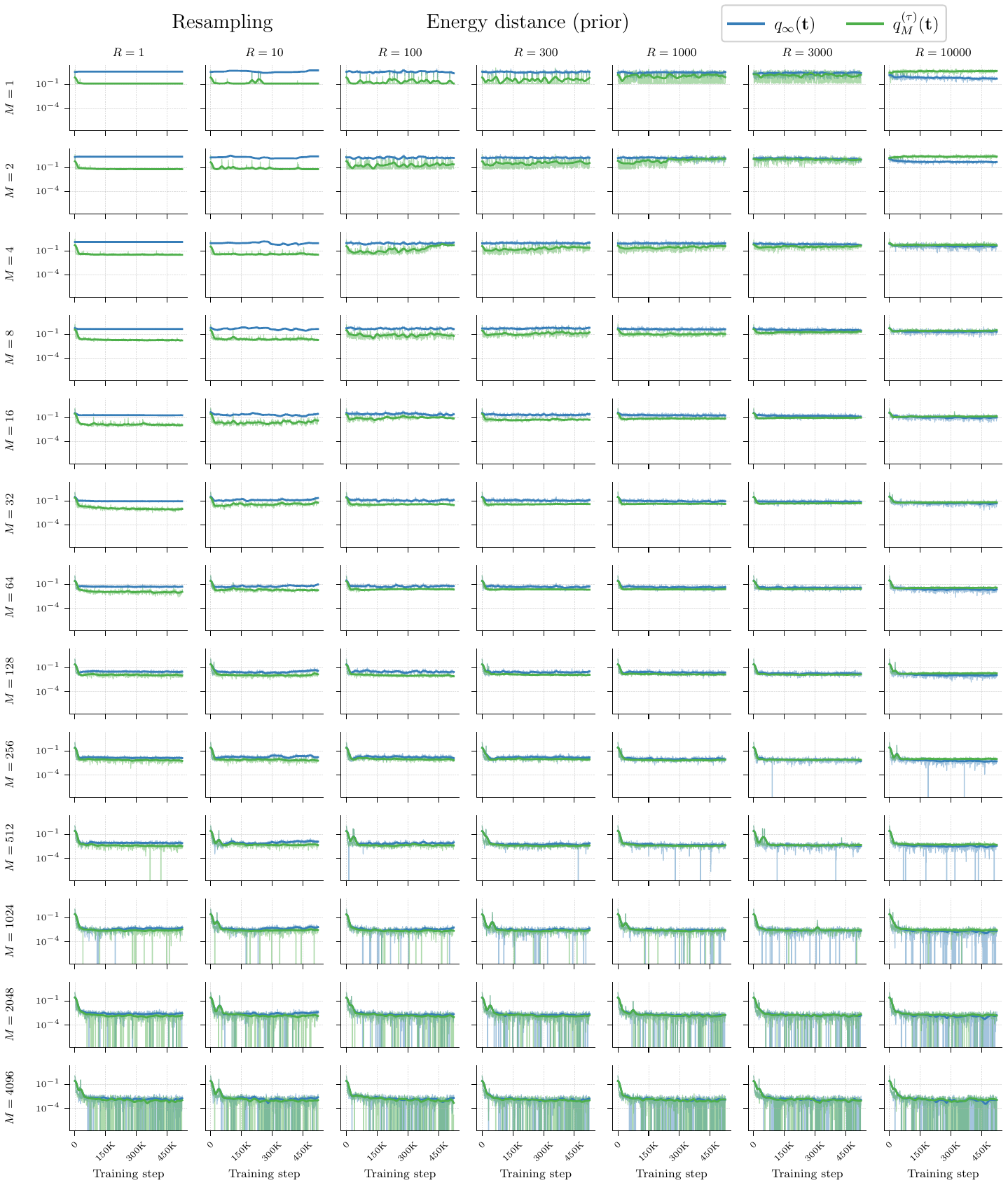}
  \caption{\textbf{Energy distance between the transformer's implicit prior and the $\text{Uniform}(\mathcal T^{(\tau)}_M)$ and $\mathcal N(0, I_D)$ priors throughout training under resampling non-stationarity.} For each combination of task diversity $M$ and sample number $R$, we use predictive Monte Carlo to extract the transformer's implicit prior over task vectors throughout training, and compare to the baseline priors via energy distance.}
  \label{fig:dirichlet-grid-energy}
\end{figure*}

\Cref{fig:mala-grid-delta-id,fig:mala-grid-delta-ood,fig:mala-grid-energy,fig:dirichlet-grid-delta-id,fig:dirichlet-grid-delta-ood,fig:dirichlet-grid-energy} show mean squared prediction differences $\Delta_\text{PT,dMMSE}$, $\Delta_\text{PT,Ridge}$ and the energy distance to the reference priors throughout training for random walk and resampling settings.
For each metric we show the full grid of training trajectories with each combination of task diversity $M$ and non-stationarity parameter ($\gamma$ for random walk, $R$ for resampling). Curves are Gaussian-smoothed with $\sigma = 3$ samples (one sample per 800 training steps). The figures show a decrease in the task diversity threshold as the degree of non-stationarity increases. For low task diversity and low degree of non-stationarity, the model can track the moving task distribution. However, as the task diversity increases, and as the degree of non-stationarity increases, the model struggles to track the moving task distribution, instead differing to the generalising approach.

\begin{figure*}[p]
  \centering
  \includegraphics[width=\textwidth,height=0.9\textheight,keepaspectratio]{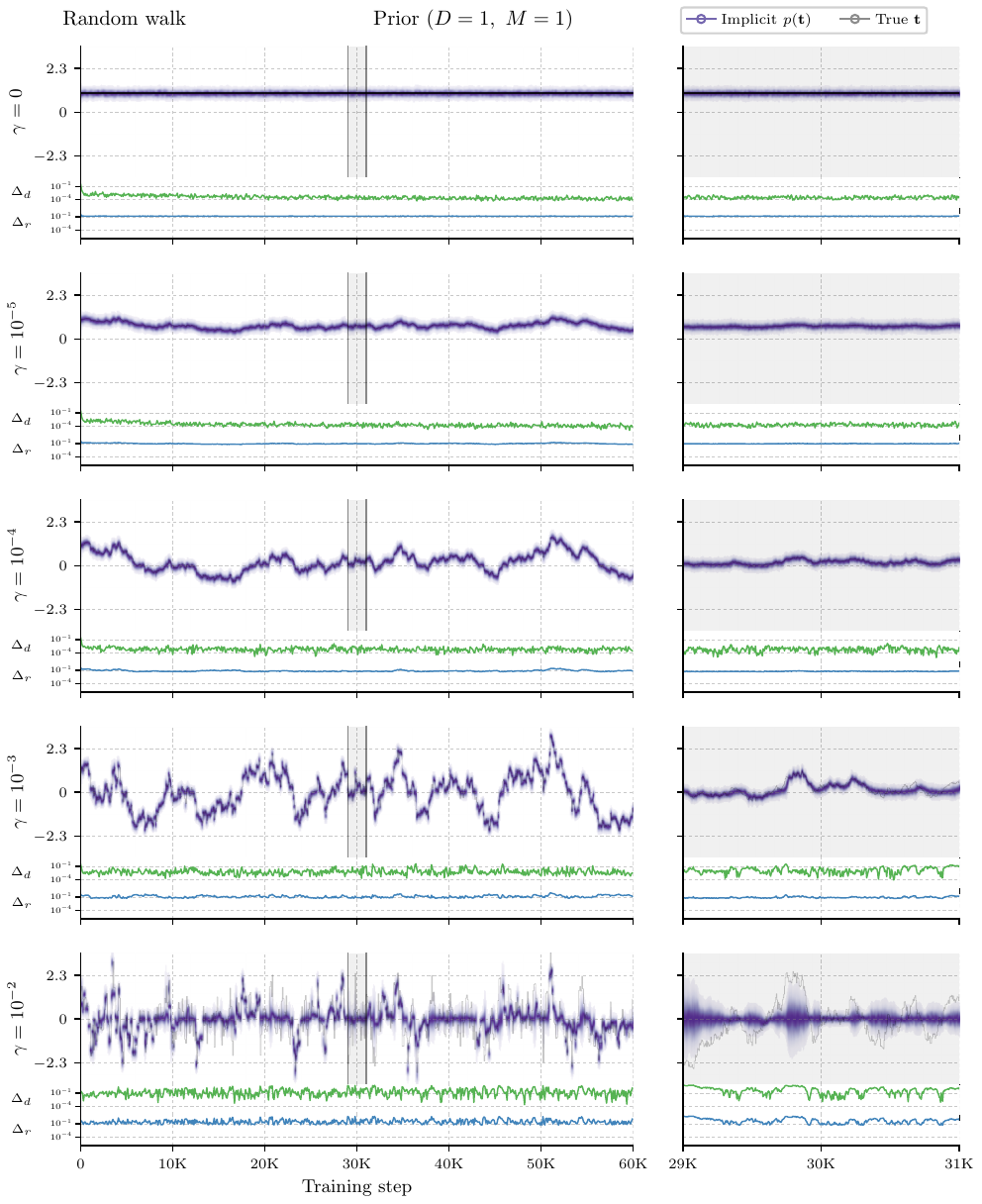}
  \caption{\textbf{Implicit prior over a 1D task vector against the true task during training, across MALA step sizes.} As in \cref{fig:prior-over-time-callout}, we use predictive Monte Carlo to extract the transformer's implicit prior over the task vector $p(\mathbf t)$ (purple) and compare it to the true task $\mathbf t$ (black), for a one-dimensional MALA setting with task dimension $D = 1$, task diversity $M = 1$ and $\gamma \in \{0, 10^{-5}, 10^{-4}, 10^{-3}, 10^{-2}\}$. On the left, we show the task vector over the initial $60$K steps of training. On the right, we zoom in on the task vectors during steps $29$K--$31$K. As $\gamma$ increases, the random walk moves faster and the implicit prior tracks it less precisely. We additionally plot $\Delta_d$ and $\Delta_r$ over each training step, the mean squared distances between the transformer and the dMMSE and ridge predictors on the training distribution.}
  \label{fig:prior-over-time-mala}
\end{figure*}

\begin{figure*}[p]
  \centering
  \includegraphics[width=\textwidth,height=0.9\textheight,keepaspectratio]{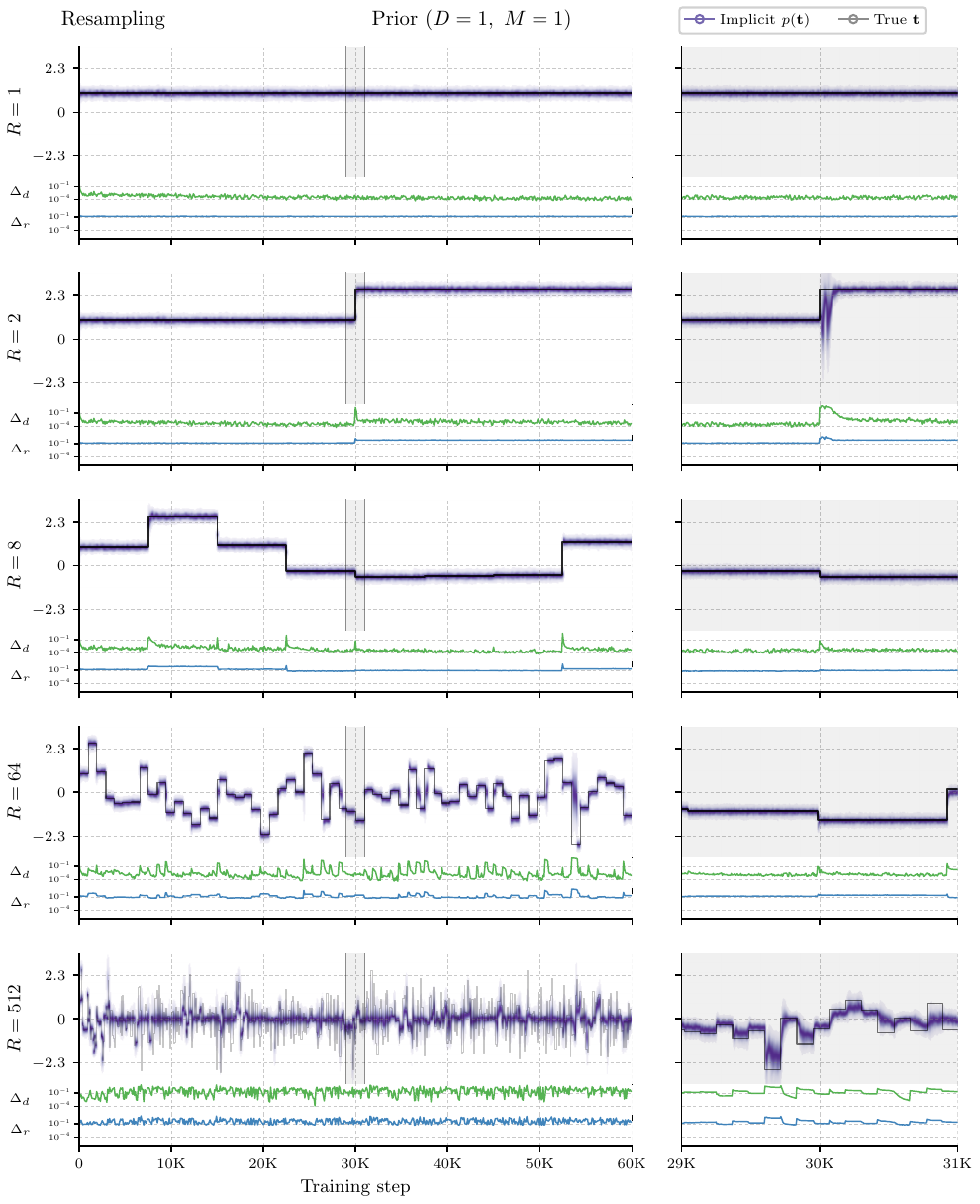}
  \caption{\textbf{Implicit prior over a 1D task vector against the true task during training, across resampling rates.} As in \cref{fig:prior-over-time-mala}, we use predictive Monte Carlo to extract the transformer's implicit prior over the task vector $p(\mathbf t)$ (purple) and compare it to the true task $\mathbf t$ (black), for a one-dimensional modified Dirichlet setting where we use $R$ independently-sampled task sets at equally spaced intervals with task dimension $D = 1$, task diversity $M = 1$ and $R \in \{1, 2, 8, 64, 512\}$. On the left, we show the task vector over the initial $60$K steps of training. On the right, we zoom in on the task vector during steps $29$K-$31$K. As $R$ increases, the task vector changes more frequently and the implicit prior tracks it less precisely. We also plot $\Delta_d$ and $\Delta_r$, the mean squared distances between the transformer and the dMMSE and ridge predictors on the training distribution.}
  \label{fig:prior-over-time-dirichlet}
\end{figure*}

\Cref{fig:prior-over-time-mala,fig:prior-over-time-dirichlet} show the implicit prior histograms over training time for the dimension $D=1$, task diversity $M=1$ setting studied in \cref{sec:pursuit:visualisation}, but for additional MALA step sizes. We also consider resampling non-stationarity (with uniformly distributed sample events, different from the Dirichlet scheme discussed in \cref{sec:experiments:dirichlet}).
We see that the implicit prior closely tracks the changing tasks in most cases, though less precisely once the non-stationarity increases.

\clearpage

\section{Training instability}
\label{app:unstable}

\begin{figure*}[!b]
  \centering
  \includegraphics[width=\textwidth]{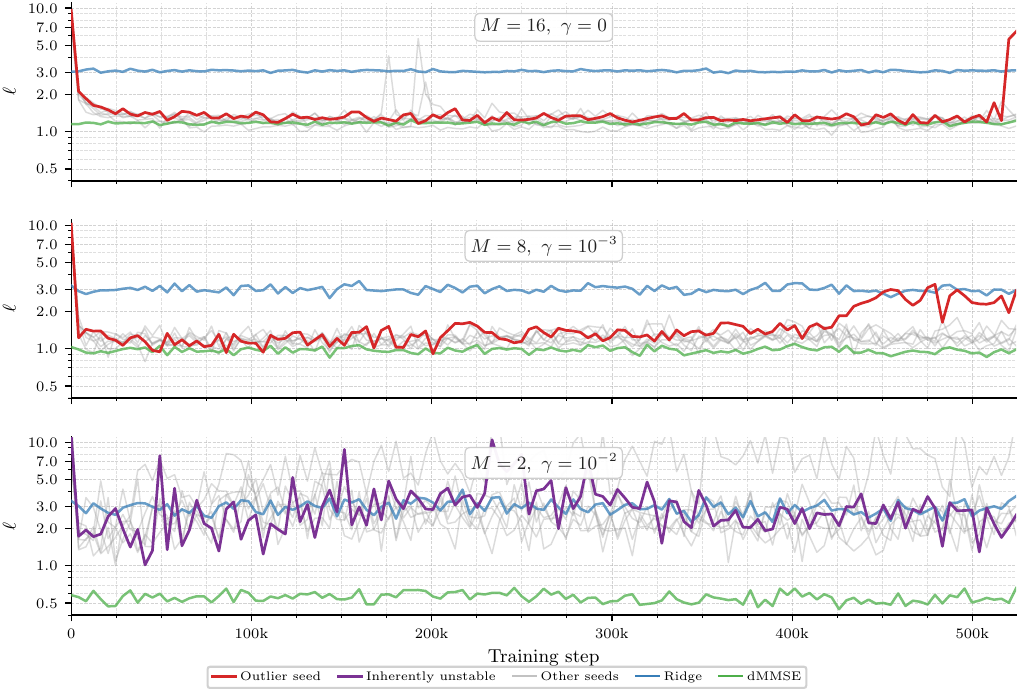}
  \caption{\textbf{Examples of training instability.} Each panel shows $\ell^M_\tau$ (log scale) against training step for a single run, evaluated every 4096 steps. In the first two panels we highlight outlier seeds (red) against the other seeds in their configuration (grey). In the third we show an entire configuration that failed to converge (purple). The losses of the ridge and dMMSE predictors are shown in blue and green respectively. The configurations are $(M = 16, \gamma = 0)$, which exhibits a late spike in loss, $(M = 8, \gamma = 10^{-3})$, which drifts upward roughly from step 400K, and $(M = 2, \gamma = 10^{-2})$, for which no seed converged.}
  \label{fig:training-instability}
\end{figure*}

In \cref{sec:experiments}, our initial sweep produced a small number of runs with training stability issues, which we summarise in \cref{fig:training-instability}. These broadly fell into two types of issues.

\begin{enumerate}
    \item
        The first type of issue was a divergence of loss compared to the other seeds. In five of our runs across the \cref{sec:experiments:mala} sweep and two runs across the \cref{sec:experiments:dirichlet} sweep, a single seed departed from the trajectory followed by the other seeds in the same configuration. The first two panels of \cref{fig:training-instability} show representative examples. The $(M = 16, \gamma = 0)$ run showed an abrupt spike in loss near the end of training, not seen in the other seeds, while the $(M = 8, \gamma = 10^{-3})$ run drifted slowly upwards starting from around step 400K. We re-ran these outlier seeds with new seeds and used the replacement data in \cref{fig:mala,fig:dirichlet}.
    \item
        The second type of instability was a configuration-wide failure to converge. In the low-task-diversity, large-step-size portion of the sweep, every seed oscillated chaotically with training MSE between roughly 1 and 10. The loss in these runs sits near that of ridge, but $\Delta_{\text{PT, ridge}}^{q_M^{(\tau)}}$ remains large, so we cannot interpret these runs as having found the ridge solution. Because no seed in this configuration converged, we did not replace these runs in \cref{fig:mala,fig:dirichlet}.
\end{enumerate}
\end{document}